\documentclass[letterpaper,journal]{IEEEtran}
\IEEEoverridecommandlockouts 

\usepackage{amsmath,amsfonts}
\usepackage{algorithmic}
\usepackage{array}
\usepackage[caption=false,font=normalsize,labelfont=sf,textfont=sf]{subfig}
\usepackage{textcomp}
\usepackage{stfloats}
\usepackage{url}
\usepackage{verbatim}
\usepackage{graphicx}
\usepackage{capt-of}
\def\BibTeX{{\rm B\kern-.05em{\sc i\kern-.025em b}\kern-.08em
    T\kern-.1667em\lower.7ex\hbox{E}\kern-.125emX}}
\usepackage{balance}

\usepackage[nohyperlinks, printonlyused, nolist]{acronym}
\begin{acronym}
\acro{AI}[AI]{Artificial Intelligence}
\acro{LLM}[LLM]{Large Language Model}
\acro{VLA}{Vision Language Action}
\acro{VLM}{Vision Language Model}
\acro{RT}{Real-Time}
\acro{GT}{Ground Truth}
\acro{GNSS}{Global Navigation Satellite System}
\acro{RTK}{Real-time Kinematic Positioning}
\acro{LiDAR}{Light Detection and Ranging}
\acro{IMU}{Inertial Measurement Unit}
\acro{RPE}{Relative Position Error}
\acro{RRE}{Relative Rotation Error}
\acro{APE}{Absolute Position Error}
\acro{ARE}{Relative Rotation Error}
\acro{RL}{reinforcement-learning}
\acro{HF}{Holistic Fusion}
\acro{MAP}{Maximum a Posteriori}
\acro{SLAM}{Simultaneous Localization and Mapping}
\acro{GUI}{Graphical User Interface}
\acro{GSL}{Gas Source Localization}
\end{acronym}

\usepackage[dvipsnames]{xcolor}
\definecolor{darkgreen}{rgb}{0.0, 0.5, 0.0}
\usepackage{array}
\usepackage{tabularx}
\newcolumntype{K}[1]{>{\RaggedRight\arraybackslash}p{#1}}
\usepackage{booktabs}   % \toprule \midrule \bottomrule
\usepackage{multirow}   % category cells spanning multiple rows
\usepackage{siunitx}    % nice units and math (optional but recommended)
\usepackage{ragged2e}    % \RaggedRight inside p-columns
\usepackage[version=4]{mhchem}
\usepackage{orcidlink}

\usepackage[nospace]{cite}

\usepackage{hyperref}
\usepackage{cleveref}
\hypersetup{pdfpagemode=UseNone}
\hypersetup{
    colorlinks=true,
    linkcolor=blue,
    linkbordercolor=white,
    filecolor=magenta,
    citecolor=cyan,
    urlcolor=blue
    }

\newcommand{\states}{\mathcal{X}}
\newcommand{\Measurements}{\mathcal{Z}}
\DeclareMathOperator*{\argmax}{arg\,max}
\DeclareSIUnit{\herz}{Hz}
\DeclareSIUnit{\ppm}{ppm}

\makeatletter
\renewcommand\subsubsection{\@startsection{subsubsection}{3}{\parindent}%
  {0em}%  no vertical space before
  {0em}%     no space after (run-in heading)
  {\normalfont\normalsize\bfseries}} % bold instead of italic
\makeatother

\newcommand\copyrighttext{%
       \scriptsize \textcopyright 2026 IEEE. Personal use of this material is permitted. Permission from IEEE must be obtained for all other uses, including reprinting/republishing this material for advertising or promotional purposes, collecting new collected works for resale or redistribution to servers or lists, or reuse of any copyrighted component of this work in other works.}
\newcommand\copyrightnotice{%
       \begin{tikzpicture}[remember picture,overlay]
               \node[anchor=south,yshift=8pt] at (current page.south) {\fbox{\parbox{\dimexpr\textwidth-\fboxsep-\fboxrule\relax}{\copyrighttext}}};
       \end{tikzpicture}%
}

\newif\ifshowappendix
\showappendixtrue
\begin{document}

\title{Large-Scale Autonomous Gas Monitoring for Volcanic Environments: \\
A Legged Robot on Mount Etna}

% Double-blind review
\newif\ifanonymous
% \anonymoustrue % Uncomment during double-blind review
\anonymousfalse % Uncomment for latex or publication

\DeclareRobustCommand{\anontext}[2]{%
  \ifanonymous
    \textcolor{orange}{[#2]}%
  \else
    #1%
  \fi
}

\author{
\anontext{
Julia Richter$^{1}$\,\orcidlink{0009-0001-5477-6440},
Turcan Tuna$^{1}$\,\orcidlink{0000-0001-8662-4890},
Manthan Patel$^{1}$\,\orcidlink{0000-0002-9171-2139},
Takahiro Miki$^{1}$\,\orcidlink{0000-0001-8556-2819},
Devon Higgins$^{2}$\,\orcidlink{0000-0003-1147-1724},
James Fox$^{2}$\,\orcidlink{0000-0000-0000-0000},\\
Cesar Cadena$^{1}$\,\orcidlink{0000-0002-2972-6011},
Andres Diaz$^{2}$\,\orcidlink{0000-0002-9628-3329}, 
and Marco Hutter$^{1}$\,\orcidlink{0000-0002-4285-4990}
\thanks{$^1$ Robotic Systems Lab (RSL), ETH Zürich, Zürich, Switzerland}
\thanks{$^2$ Intelligent Sensor Solutions, INFICON Inc., East Syracuse, NY, US}
\thanks{\textit{The supplementary video and raw scientific data are available on our webpage: \url{https://leggedrobotics.github.io/etna-expedition/}}}
}{anonymous authors\thanks{\textit{Supplementary video: \url{https://youtu.be/Gk317LnNqTs}}}}
}

\IEEEaftertitletext{%
  \begin{center}
    \includegraphics[width=.9\textwidth]{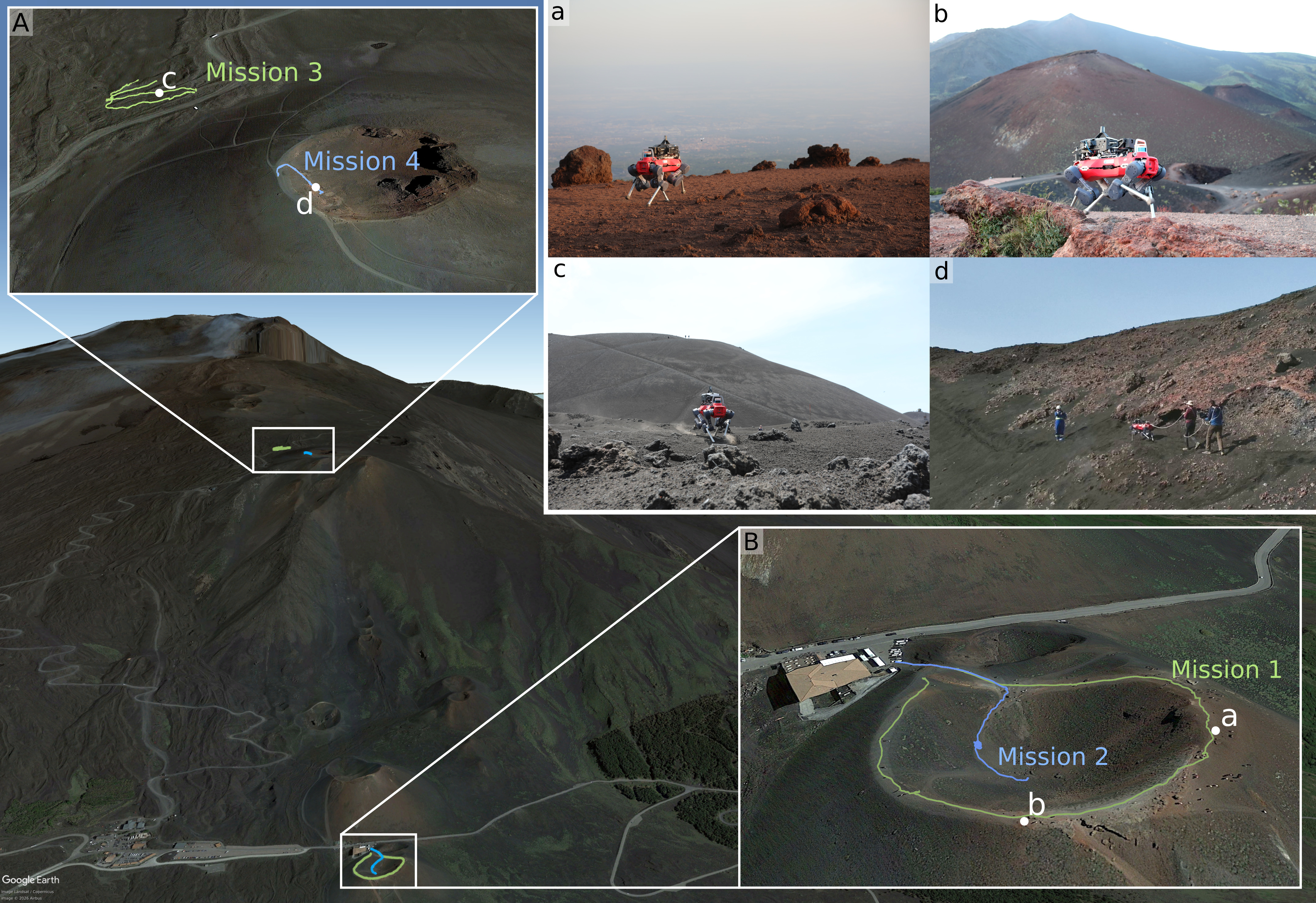}
    \captionof{figure}{Aerial overview of Mount Etna with the four completed missions indicated. (A--B) Zoomed views showing detailed trajectories of two mission areas. (a--c) Images from autonomous helium-mapping missions. (d) Image from the teleoperated mission measuring volcanic gases at an active crater.}
    \label{fig:conceptual_scenario}
  \end{center}
  \vspace{0.5\baselineskip}% (optional: adjust spacing before Abstract)
}

\maketitle

%%%%%%%%%%%%%%%%%%%%%%%%%%%%%%%%%%%%%%%%%%%%%%%%%%%%%%%%%%%%%%%%%%%%%%%%%%%%%%%%
\begin{abstract}
\noindent Volcanic gas emissions are key precursors of eruptive activity.
Yet, obtaining accurate near-surface measurements remains hazardous and logistically challenging, motivating the need for autonomous solutions.
Limited mobility in rough volcanic terrain has prevented wheeled systems from performing reliable in~situ gas measurements, reducing their usefulness as sensing platforms.
We present a legged robotic system for autonomous volcanic gas analysis, utilizing the quadruped ANYmal, equipped with a quadrupole mass spectrometer system.
Our modular autonomy stack integrates a mission planning interface, global planner, localization framework, and terrain-aware local navigation.
We evaluated the system on Mount Etna across three autonomous missions in varied terrain, achieving successful gas-source detections with autonomy rates of 93–100\%.
In addition, we conducted a teleoperated mission in which the robot measured natural fumaroles, detecting sulfur dioxide and carbon dioxide.
We discuss lessons learned from the gas-analysis and autonomy perspectives, emphasizing the need for adaptive sensing strategies, tighter integration of global and local planning, and improved hardware design.

\end{abstract}

\unless\ifanonymous
    \copyrightnotice
\fi

%%%%%%%%%%%%%%%%%%%%%%%%%%%%%%%%%%%%%%%%%%%%%%%%%%%%%%%%%%%%%%%%%%%%%%%%%%%%%%%%
\section{INTRODUCTION}
\label{sec:INTRODUCTION}

About 29 million people live within \SI{10}{\kilo\meter} of an active volcano, with fatalities recorded in all but two years between 1997 and 2017~\cite{brown2017volcanic}.
Scientists have been focusing on protecting affected populations through monitoring and analysis that enable timely warnings and evacuation.
Monitoring is often conducted through the deployment of sensor networks that track ground deformation, seismic activity, and volcanic gas emissions. Since eruptions are typically preceded by changes in these signals on timescales from years to minutes, such measurements can enable forecasts of eruption timing~\cite{acocella2024towards}.
These signals can be acquired remotely or in~situ, where these different data collection schemes are complementary to each other~\cite{acocella2024towards}.
This work focuses on autonomous in~situ gas measurements, which enable the real-time estimation of volatile concentration and composition relevant to magmatic processes~\cite{carn2015gas, diaz2002mass}.

Conducting in~situ gas measurements requires bringing instruments to fumaroles – small openings and fractures that emit volcanic gases and steam and typically cluster near active vents. As a consequence, these sites are often hazardous due to unstable terrain, toxic gases (e.g., \ce{SO2}, \ce{H2S}), and sudden explosive phreatic or pyroclastic eruptions, making human-assisted sampling particularly dangerous. \looseness-1
Mobile robotic systems enable event-driven sampling in demanding and evolving volcanic environments. They can simultaneously acquire gas, thermal, and topographic data, providing essential context for interpretation~\cite{diefenbach2024special,james2020volcanological,muscato2012volcanic}.
Unlike fixed sensor stations, which remain in place until a hazardous human visit repositions them, a robot can be redeployed remotely to follow shifting plume activity, dispatched rapidly to emergent degassing, and used to survey large areas with a single high-performance instrument.

Mobility serves two distinct roles for gas sensing: reaching predetermined emission sites to perform measurements, and reactively searching for unknown sources by interpreting concentration gradients in real time. 
The latter is the domain of \ac{GSL}, where techniques such as probabilistic source estimation and informative path planning drive a robot's trajectory based on live gas and wind readings~\cite{francis2022gas}. However, existing \ac{GSL} methods rely on dispersion models that assume moderate, quasi-steady airflow and have been validated predominantly in controlled indoor or small-scale outdoor settings~\cite{francis2022gas,li2011odor}. The strong, variable winds encountered on volcanic summits fall well outside these validated conditions. 
Accordingly, robotic volcanic gas sensing has relied on the first role: reaching known or suspected emission sites and recording gas measurements along the trajectory.

Prior work has pursued this approach with unmanned aerial vehicles (UAVs) for plume sampling~\cite{james2020volcanological,diefenbach2024special,rolland2024autonomous}. 
UAVs are effective for vent-level plumes because they can approach vent airspace while maintaining a safe standoff. Near-surface degassing, however, occurs at fumaroles distributed across the volcano with variable flux and composition (one example shown in Figure \ref{fig:plume_viz}). Rapid dilution means that accurate sampling requires proximity to the source~\cite{carn2015gas}, which is particularly challenging for UAVs in complex terrain and unstable winds. 

Different research groups have developed specialized unmanned ground vehicles (UGVs) to address the challenges of source-proximal plume measurements~\cite{muscato2012volcanic,bares1999dante}.
However, this also introduces locomotion and terrain hazards, such as loose tephra, steep slopes, lava ridges, and boulder fields~\cite{muscato2012volcanic}. 
Current wheeled and wheeled-legged rovers for volcanic monitoring have limited adaptability to uneven ground, compromising their stability. To protect sensitive instruments and avoid rollover, they must move slowly, which reduces operational efficiency and limits their practicality as in~situ measurement platforms.
Additionally, although several robotic systems have been developed for volcanic gas measurement, to the best of the authors' knowledge, no ground-based robot has conducted successful in~situ measurements on an active volcano. The primary obstacle lies in the inaccessibility of fumaroles, which demand advanced locomotion capabilities and sufficient efficiency to reach remote sites.
Moreover, current ground systems are primarily teleoperated; however, radio links can become intermittent during eruptions or in rugged terrain, making continuous teleoperation impractical and motivating the development of autonomous mission capabilities~\cite{usgs2015comms}.

In recent years, legged robots have emerged as a new class of ground platforms capable of traversing diverse terrains. 
Their feasibility for in-the-wild applications has been demonstrated in increasingly challenging outdoor settings.
Existing deployments in natural environments have predominantly relied on teleoperation, e.g., for wildlife observation~\cite{melo2023animal} or soil sampling~\cite{wilson2021spatially}, while a few works incorporate partial autonomy, e.g., for planetary analog exploration~\cite{arm2023scientific} or forest inventory~\cite{mattamala2024autonomous}.
In these systems, the mission design remains human-in-the-loop: the operator takes on the role of the global planner by specifying a dense sequence of waypoints, placed close enough that only local obstacle avoidance is required between them. The robot then follows these waypoints under safety supervision~\cite{arm2023scientific,mattamala2024autonomous,wilson2021spatially}.
Notably, no legged robot has been deployed autonomously in a volcanic environment, where GNSS reception is unreliable, geometric structure is sparse, and terrain conditions change continuously.

In this paper, we present a legged robotic system for autonomous volcanic gas monitoring that combines a mobile mass spectrometer with a high-level autonomy stack.
We detail the integration of the mass spectrometer, including the design decisions required to maintain vacuum stability and instrument protection during locomotion in rough terrain.
We describe the autonomy stack developed for long-range navigation in GNSS-degraded volcanic terrain, including an intuitive mission planning interface, a multi-sensor fusion state estimator that operates without RTK-GNSS corrections, and a curvature-adaptive lookahead scheme for terrain-aware local navigation.
The system is validated through three autonomous field missions on Mount Etna, representing the first successful volcanic gas measurements performed by a ground robot in an active volcanic area.
We further discuss key lessons learned from these real-world deployments, highlighting practical challenges and their implications for future autonomous systems for gas sensing and long-range navigation in natural environments.
Ultimately, this work aims to support volcanologists in both regular and event-driven monitoring of active areas by enabling source-proximal gas measurements beyond the reach of human operators. \looseness-1

\section{SYSTEM DESCRIPTION}
\label{sec:SYSTEM DESCRIPTION}
\subsection{Robotic Platform}

\begin{figure}[t]
    \centering
    \includegraphics[width=\linewidth]{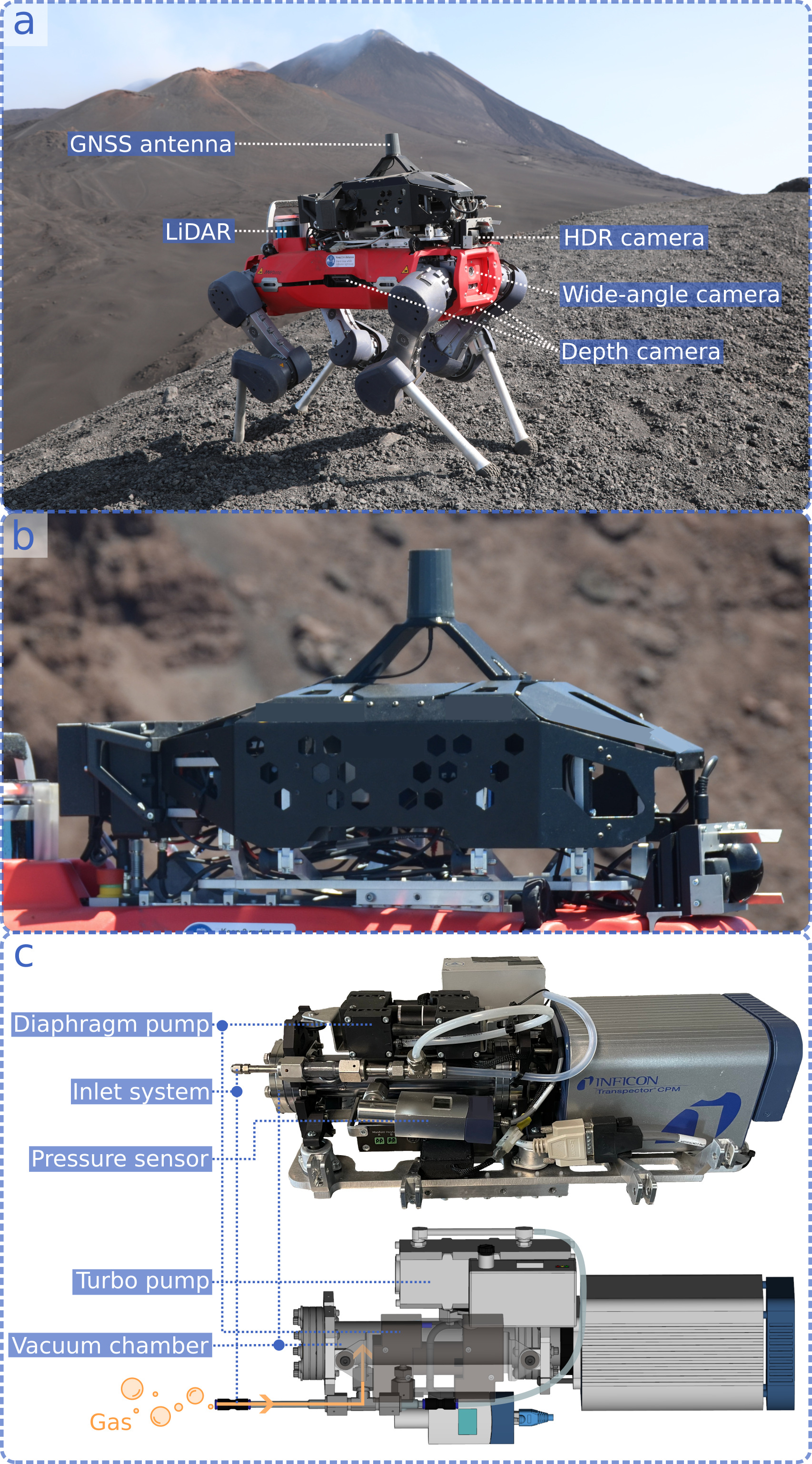}
    \caption{Overview of the robotic system equipped with the gas-sensing payload and perception sensors.
    (a) Robot with labeled sensor components. 
    (b) Gas-sensing payload with rollcage.
    (c) Transpector\textsuperscript{\textregistered} MPH quadrupole mass spectrometer subsystem.}
    \label{fig:hardware_setup}
\end{figure}

Our robotic system consists of two parts, namely, the locomotion platform ANYmal, an off-the-shelf quadruped produced by ANYbotics AG, and a gas-sensing payload.
The robot weighs \SI{50}{\kilogram}, carries up to \SI{12}{\kilogram} of payload, and runs for about \SI{1}{\hour} at maximum payload.
As labeled in Figure~\ref{fig:hardware_setup}a, the robot carries six RealSense D435i depth cameras for close-range perception, a Velodyne VLP-16 LiDAR for long-range measurements, an IMU, and a Swift Navigation Piksi Multi \ac{GNSS} receiver operating without Real-Time Kinematic (RTK) corrections. Wide-angle RGB cameras provide situational awareness for the operator. Computation runs on two internal Intel Core i7 processors and an NVIDIA Jetson Orin NX, the latter used for GPU-accelerated elevation mapping (Section~\ref{subsec:elevation_mapping}).

\begin{figure*}[t]
    \centering
    \includegraphics[width=.9\textwidth]{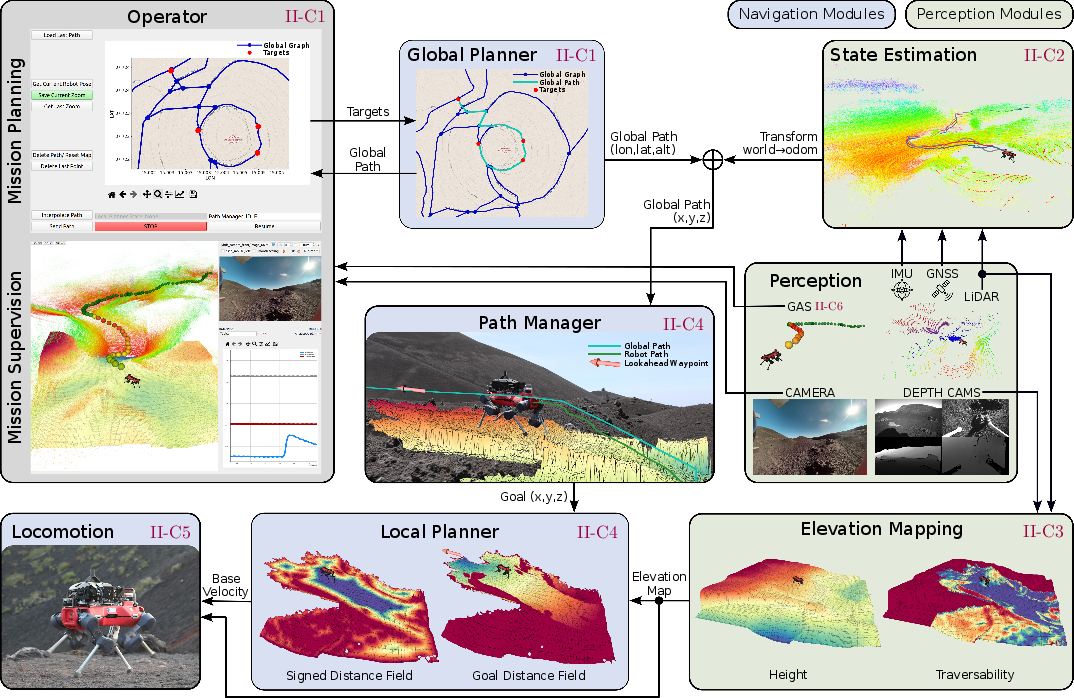}
    \caption{
Overview of software architecture.
The architecture is organized into two main parts: the \textcolor[HTML]{4065c0}{blue} navigation modules, responsible for global and local path planning and locomotion control, and the \textcolor[HTML]{759551}{green} perception modules, which handle state estimation and elevation mapping.
Information flows from high-level mission planning (top left) to low-level actuation (bottom left), while feedback from perception enables autonomous operation and live supervision of mapping and gas measurements.
The numbers in \textcolor[HTML]{9d0141}{pink} correspond to the sections of this paper in which the respective modules are described.
}
\label{fig:software_architecture}
\end{figure*}

\subsection{Scientific Gas-Sensing Payload}

To analyze volcanic gases in~situ, the sensing system must provide precise, selective, and robust measurements under harsh environmental conditions. 
Conventional approaches typically employ non-dispersive infrared (NDIR) sensors (e.g., for \ce{CO2}, \ce{CO}, \ce{H2O}), electrochemical sensors (e.g., for \ce{SO2}, \ce{H2S}, \ce{HCl}, \ce{HF}), and semiconductor sensors (e.g., for \ce{H2})~\cite{carn2015gas}. 
While effective for targeted gas components, these sensors have a limited dynamic range and are prone to cross-sensitivity and degradation in acidic volcanic environments~\cite{carn2015gas}.
To address these limitations, we employ a compact mass spectrometer as the primary sensing instrument, which has been previously validated in volcanic environments~\cite{diaz2015unmanned}. 
This analytical instrument identifies compounds based on their mass-to-charge ratio, providing high selectivity and accuracy across a wide range of gas components~\cite{diaz2002mass}. 
In practice, it can resolve characteristic atomic mass units (amu) of various compounds, such as hydrogen (\ce{H2}, 2 amu), helium (\ce{He}, 4 amu), carbon dioxide (\ce{CO2}, 44 amu), and sulfur dioxide (\ce{SO2}, 64 amu), along with many other gases, isotopes, and molecular fragments.
Moreover, its six-order-of-magnitude dynamic range prevents saturation at high concentrations, enabling reliable measurements both within dense fumarolic plumes and in downwind regions with sub-ppm (parts per million) concentrations.

The instrument we employ is an INFICON Transpector\textsuperscript{\textregistered} MPH quadrupole mass spectrometer, which has a mass range of \SI[parse-numbers = false]{1-200}{amu} and a detection limit of \SI{<1}{ppm} in controlled environments.
Operation requires a high-vacuum environment, which is provided by a compact two-stage pumping setup consisting of a miniature diaphragm pump and a very small turbomolecular pump (Figure~\ref{fig:hardware_setup}c).
Ambient air enters the system through a foam inlet filter and a \SI{1}{\meter} stainless-steel capillary with a fixed flow restriction, which controls the gas flow into the measurement chamber and determines the response time of approximately \SI{3}{\second}.

The complete mass spectrometer subsystem weighs \SI{8.7}{\kilogram}. 
As shown in Figure~\ref{fig:hardware_setup}b, we constructed an aluminum rollcage weighing approximately \SI{3}{\kilogram} to protect the instrument from fall damage. 
The mass spectrometer is powered through the robot's output port, with an average power consumption of \SI{36}{\watt}, and communicates via TCP/IP through its native API.
 
\subsection{Software Architecture}

The software architecture, shown in Figure~\ref{fig:software_architecture}, follows a modular design structured into two main categories. 
\textcolor[HTML]{4065C0}{Blue} denotes the navigation modules, responsible for global and local path planning and for executing velocity commands through the locomotion controller. 
\textcolor[HTML]{759551}{Green} indicates the perception modules, which process data from multiple onboard sensors to perform state estimation and elevation mapping. 
Together, these components enable the hierarchical planning and control pipeline described below, which orchestrates the robot's autonomous behavior from high-level mission specification down to low-level actuation.
At the highest level, a \textit{Mission Planning} \ac{GUI} allows the \textit{Operator} to define mission \textcolor[HTML]{FF0000}{targets}, which serve as targets for the \textit{Global Planner} to generate a global \textcolor[HTML]{10C7B8}{path}. 
Since this path is represented in geodetic coordinates (latitude, longitude, and altitude), it is transformed into the robot's local reference frame through sensor fusion of global \ac{GNSS} data with point-cloud \ac{SLAM} and kinematic measurements.
From the transformed \textcolor[HTML]{10C7B8}{path}, a \textcolor[HTML]{E36C5E}{lookahead waypoint} is sampled and passed to the \textit{Local Planner}. 
The \textit{Local Planner} then computes a feasible local trajectory toward the \textcolor[HTML]{E36C5E}{lookahead waypoint} based on the height and traversability information provided by the \textit{Elevation Mapping} module. 
Finally, the resulting base velocity commands are executed by a \ac{RL}  \textit{Locomotion} controller that drives the individual leg actuators to achieve the desired body motion.
During the mission, the operator can monitor the system through the \textit{Mission Supervision} interface, which provides access to the assembled SLAM map, live camera feeds, and real-time gas measurement data.

\subsubsection{Mission Planning}
\label{subsec:mission_planning}

Since the robot must cover long distances, the autonomy stack requires a dedicated global path planning module.
Rather than relying on a fully automated planner, we developed an intuitive operator interface that keeps the human in the loop while simplifying the planning process.
The interface displays globally referenced satellite data in geodetic latitude-longitude coordinates, providing the operator with sufficient environmental context for planning.
Ideally, this visualization is based on high-resolution satellite imagery.
Since such data is not universally available, we default to the Tracestrack Topo layer of OpenStreetMap\footnote{\url{https://www.openstreetmap.org/\#layers=P}}, which includes path networks, types, and dense contour lines that allow recognizing topographic features such as craters. 
The data is fetched via the OpenStreetMap API for a user-specified area, and an example is shown in Figure~\ref{fig:software_architecture} (\textit{Global Planner}). 

To plan a mission, the operator translates mission objectives into a sequence of latitude-longitude points at arbitrary spacing by clicking directly on the shown image.
The global planner then computes a path between these targets.
For this, it operates on a navigation graph generated offline from OpenStreetMap data using the OSMnx Python wrapper~\cite{boeing2025modeling}, visualized in \textcolor[HTML]{0005c1}{blue} in Figure~\ref{fig:software_architecture}. On this graph, an A* search determines the cost-optimal global path, where the cost between two connected nodes is their Euclidean distance.
Although slope and surface characteristics naturally influence traversal cost, publicly available digital elevation models have coarse resolutions (\SI[parse-numbers = false]{10-30}{\meter}), making it difficult to estimate these factors a priori.
Targets located outside the graph are linearly connected to their nearest projection point.
As shown in Figure~\ref{fig:software_architecture}, the mission planning \ac{GUI} is used by the operator and outputs a global path in latitude-longitude-altitude coordinates, which is transformed into the Cartesian robot frame based on the state estimation described in the following section.

\subsubsection{State Estimation}
\label{subsec:state_estimation}

Accurate state estimation, including a reliable world-referenced robot pose, is a prerequisite for executing long-range and large-scale navigation tasks.
We build on the optimization-based multi-sensor fusion framework \ac{HF}~\cite{nubert2026holistic} and adapt it to better handle the challenges of long-range deployments in featureless volcanic environments, such as numerical ill-conditioning, intermittent and unreliable \ac{GNSS}, and inconsistent estimates that drift over time. The underlying factor-graph formulation is given in the supplementary material.
It fuses three complementary sources: \textit{i)} globally referenced but noisy non-RTK \ac{GNSS} measurements, \textit{ii)} globally consistent but drifting \ac{LiDAR}-SLAM poses, and \textit{iii)} short-term stable odometry from ANYbotics' internal kinematic-inertial estimator. 
We decided to utilize a LiDAR-SLAM module rather than LiDAR odometry since GNSS availability on Mount Etna was uncertain before deployment, making a SLAM-based solution the more robust choice. In addition, SLAM provides a consistent scene representation and a fixed-world frame after geo-localization, enabling reliable localization against a previously constructed submap in the feature-poor volcanic terrain.

For \ac{LiDAR}-SLAM poses, we employ the degeneracy-aware X-ICP~\cite{xicp} \ac{LiDAR}-based SLAM framework. 
This module provides smooth, temporally consistent pose estimates within the evolving \ac{LiDAR} map, but its accuracy ultimately depends on the observability of the incoming \ac{LiDAR} data against the map.
In loose soil, the robot's feet slip, causing sudden accelerations that, in turn, introduce motion distortions into the \ac{LiDAR} observations.
Combined with the already sparse geometric structure typical of volcanic terrain, these distortions can cause \ac{LiDAR}-SLAM to drift over long distances.

We make two central design choices in order to enhance \ac{HF} and make it more robust for long-distance deployment.
First, instead of treating the \ac{LiDAR}-SLAM output as incremental odometry, we introduce \emph{absolute} \ac{LiDAR}-SLAM pose factors into the graph.
Second, we explicitly estimate the alignment between the evolving \ac{LiDAR}-SLAM map and the world frame as part of the optimization.
Concretely, we estimate the time-varying rigid-body transform between the LiDAR-SLAM map frame and the world frame at regular intervals. The change between sequential rigid-body transforms is thereby modeled as a random walk.
The resulting estimator remains stable during \ac{GNSS} outages, and quickly re-anchors to the global frame once absolute information is restored.

This variable anchors \ac{LiDAR}-SLAM poses to the global estimate on the fly, enabling the fusion to (a) maintain smooth, globally consistent trajectories across \ac{GNSS} dropouts, (b) query the robot state directly in the map frame at any time, and (c) model uncertainty growth according to the assumed drift characteristics of the mapping back end (rather than just local registration noise). 
Figure~\ref{fig:software_architecture} illustrates how the \textit{State Estimation} module interfaces with the navigation stack.
Its main task is to localize the robot in the world frame \(\mathtt{W}\) and provide a transform between the Cartesian \(\mathtt{W}\) and the geodetic frame \(\mathtt{LLA}\) (latitude-longitude-altitude), where the global path is defined.
For this, \(\mathtt{W}\) is defined as a Cartesian East-North-Up frame on the WGS-84 ellipsoid, which is anchored at the robot's initial geographic position.
We additionally employ a geoid undulation model (EGM2008) to convert the ellipsoidal heights to orthometric heights referenced to mean sea level, thereby enhancing the accuracy of the transformation.
\looseness-1

\subsubsection{Elevation Mapping and Traversability Estimation}
\label{subsec:elevation_mapping}

Navigating in unstructured environments requires a reliable spatial representation of the surrounding terrain. 
We represent the terrain as a 2.5D elevation map, where each grid cell stores height and traversability information used by both the locomotion controller and the local planner.
We employ \textit{Elevation Mapping CuPy}~\cite{miki2022elevation}, an elevation mapping framework accelerated on the GPU for fast and efficient fusion of multi-modal sensory inputs.
As shown in Figure~\ref{fig:software_architecture}, it integrates depth data from six RGB-D cameras with measurements from the LiDAR to construct an ego-centric elevation map of size \SI{12}{\meter}$\times$\SI{12}{\meter} at a resolution of \SI{0.06}{\meter}. 
For geometric traversability estimation, we use a lightweight convolutional neural network trained on a small set of manually annotated terrain samples to predict a traversability score between 0 and 1. 
To improve prediction quality, we extend the pipeline with two additional post-processing steps to the traversability layer. First, an \textit{inpainting} operation fills gaps to ensure spatial consistency. Second, a \textit{dilation} with a kernel size of \SI{0.24}{\meter} expands non-traversable regions to increase safety margins. Both steps are motivated by the inherent uncertainty of volcanic terrain, where loose edges and irregular surfaces can compromise footholds.
Examples of the resulting maps are shown in Figure~\ref{fig:software_architecture} (\textit{Elevation Mapping}). 
The two generated layers, \textit{Height} and \textit{Traversability}, are subsequently used by the \textit{Local Planner}, described in the following section.\looseness-1

\subsubsection{Local Navigation}
\label{subsec:local_nav}

While the global planner computes a feasible high-level path, local uncertainties, such as \ac{GNSS} drift or previously unseen obstacles, increase the demand for a reactive local planner to follow this path safely.
Since local planners optimize paths primarily within the robot's field of view, we propose a curvature-adaptive lookahead sampling scheme that provides an intermediate waypoint to guide the robot along the global path. This \textit{lookahead waypoint} is \mbox{visualized} in Figure~\ref{fig:software_architecture} (\textit{Path Manager}) and updated at \SI{0.5}{\herz}.
The lookahead distance trades anticipation against path adherence: too large a value causes premature deviation near corners, while too small a value makes the robot sensitive to artifacts in the global path. We therefore adjust it to the curvature of the global path, using a longer lookahead in straight, open areas and a shorter one in tight turns.
We estimate the local path curvature from the angle between the forward and backward secant vectors at the robot's projection onto the path, and map it monotonically to a lookahead distance bounded by $L_\mathrm{min}$ and $L_\mathrm{max}$. The lookahead waypoint is obtained as an arc-length-weighted average of samples ahead of the anchor point, emphasizing nearby geometry. The full derivation is given in the supplementary material.

To follow this lookahead waypoint, we employ the \textit{Field Local Planner} previously introduced by~\cite{mattamala2022efficient}, in combination with the \textit{Traversability} layer of our elevation map. The planner fuses multiple vector fields using Riemannian Motion Policies (RMPs) for continuous obstacle avoidance and goal attraction.
For obstacle avoidance, a Signed Distance Field (SDF) is used, which generates a repulsive field that drives the robot away from obstacles.
The goal attraction is implemented as a Geodesic Distance Field (GDF), which produces an attractive field that guides the local planner toward the goal.
In unstructured outdoor terrain, noisy traversability estimates can produce disconnected traversable regions that render certain local goals unreachable within the elevation map. To prevent the local planner from pursuing such infeasible targets, we take a conservative approach and extend the GDF computation to detect unreachable goals and relocate them to the nearest reachable position.
Examples of both fields are shown in Figure~\ref{fig:software_architecture} (\textit{Local Planner}).\looseness-1

\subsubsection{Locomotion Controller}
\label{subsec:locomotion}

Locomotion is a key component of the autonomy framework, ensuring that high-level planning and sensing objectives can be executed reliably in rough volcanic terrain. 
The \ac{RL}-based controller receives base velocity commands from the \textit{Local Planner} and integrates elevation-map information for perceptive locomotion, generating coordinated leg motions that maintain stability and energy efficiency on uneven terrain.
Following the methodology of~\cite{miki2022learning}, the controller is trained entirely in simulation using a teacher-student distillation pipe\-line, with no modifications to the terrain curriculum. 
While a general policy trained with standard domain randomization can tolerate moderate payload changes, it exhibits less efficient gaits under significantly increased weight and shifted center of mass. 
To improve stability and efficiency, we trained an extended policy using payload-specific domain randomization. 
During training, the simulated robot's base mass was augmented to include the payload, and its center of mass was randomly varied within a plausible range to account for the uncertainty of the real instrument and rollcage.
This payload information was provided to the teacher policy as a privileged input, enabling it to develop locomotion strategies specifically adapted to the altered dynamics. 
The resulting student policy retained this robustness while remaining fully deployable without privileged sensing.
In practice, this approach produced a stable, terrain-adaptive gait that maintained reliable performance throughout the field campaign. 
Although the learned policy supports walking speeds of up to \SI{1.5}{\meter\per\second} and slope climbing up to \SI{25}{\degree}, a conservative limit of \SI{0.8}{\meter\per\second} was imposed during field operation for safety. \looseness-1

\subsubsection{Gas Sensing and Data Processing}

To enable autonomous gas sensing during locomotion, we developed a dedicated ROS package to interface directly with the mass spectrometer via its native API, providing full instrument control and real-time data acquisition on board the robot.
The instrument \mbox{supports two} measurement modes: a \textit{bin} scan, which repeatedly measures a predefined set of selected atomic mass values (e.g., [2, 4, 8, 44]~amu), and an \textit{analog} scan, which continuously records spectra over a specified range (e.g., 1--50~amu).
We employ the \textit{bin} mode for its higher sampling rate, which enables temporal mapping of gas concentrations during motion.
During the missions, the mass spectrometer sampled continuously at \SI{1}{\hertz}.

Through the ROS integration, all spectrometer measurements are automatically time-synchronized with the robot. 
This allows each gas sample to be spatially referenced in both global and local frames, and visualized in the mission planning interface as colored markers along the robot's trajectory (Figure~\ref{fig:software_architecture} \textit{Operator}). 
The concentration levels are also plotted in real time, providing the operator with immediate feedback for situational awareness and plume analysis.
Together, the hardware and software integration form a self-contained gas-sensing pipeline that operates autonomously alongside the navigation stack, requiring no intervention during missions.

\section{AUTONOMOUS MISSIONS}
\label{sec:MISSIONS}
\subsection{Testing Setup}
To evaluate the performance of the proposed system, field experiments were conducted on Mount Etna, Sicily, Italy.
Mount Etna is one of the most active volcanoes in the world, featuring diverse terrain ranging from solidified lava flows to loose volcanic deserts.
Our field campaign had two main objectives: measuring volcanic gases and demonstrating autonomous operation on volcanic terrain. 
The Laghetto crater had active fumaroles at the time of our tests and was therefore selected as the gas-measurement site. The path to the fumaroles, however, was narrow, steep, and covered in loose sand (Figure~\ref{fig:plume_viz}), so the gas-measurement mission was teleoperated to ensure safety. Results are discussed in Section~\ref{sec:PLUME MEASURING}.

The autonomy tests were instead performed at locations that provided safe conditions for autonomous navigation, where artificial gas sources were deployed to evaluate sensing performance.
We designed three representative scenarios that reflect typical challenges for autonomous operation on a volcano, distributed across different regions of Mount Etna, as shown in the region overview in Figure~\ref{fig:conceptual_scenario}. 
Mission~1 (Figure~\ref{fig:mission_overview}-M1) followed the rim of the Silvestri crater, featuring a long traverse with few visual landmarks for localization.
Mission~2 (Figure~\ref{fig:mission_overview}-M2) descended into the same crater, posing locomotion challenges due to steep slopes and a floor covered in rocks of different sizes.
Mission~3 (Figure~\ref{fig:mission_overview}-M3) took place in a volcanic desert near the Laghetto crater, where fine sand, numerous rocks, and degraded \ac{GNSS} reception made both locomotion and localization difficult.
During operation, data was streamed to the operator's computer for monitoring (Figure~\ref{fig:software_architecture} \textit{Mission Supervision}), while the operator remained nearby to allow for safety-critical intervention.
To demonstrate the gas-sensing capability in a controlled yet representative way, we placed helium bottles along the planned trajectories.
Helium was selected because it is inert, harmless, and easily detectable against its negligible atmospheric background.
Results of the autonomous missions are presented below.

\begin{table}[b]
\centering
\caption{Performance metrics for three autonomous missions.}
\label{tab:mission_metrics}
\footnotesize
\setlength{\tabcolsep}{5pt}
\renewcommand{\arraystretch}{1.15}
\begin{tabular}{@{}lcccc@{}}
\toprule
 & \textbf{M1} & \textbf{M2} & \textbf{M3} & \textbf{Overall} \\
\midrule
Length [\si{\meter}]        & 388    & 250  & 270    & 303    \\
Duration [mm:ss]            & 10:02  & 07:29 & 09:56   & 09:09   \\
Gas sources detected        & 3/5    & 1/1  & 1/2    & 5/8    \\
Interventions               & 4      & 0    & 3      & 2.3    \\
Intervention time [mm:ss]   & 00:43   & 00:00 & 00:22   & 00:22   \\
RAD                         & 0.0887 & 0    & 0.0490 & 0.0459 \\
AR [\%]                     & 92.8   & 100  & 96.3   & 96.4   \\
\bottomrule
\end{tabular}
\end{table}

{
\setlength{\abovecaptionskip}{2pt}
\begin{figure*}[p]
    \centering
    \includegraphics[width=.99\textwidth]{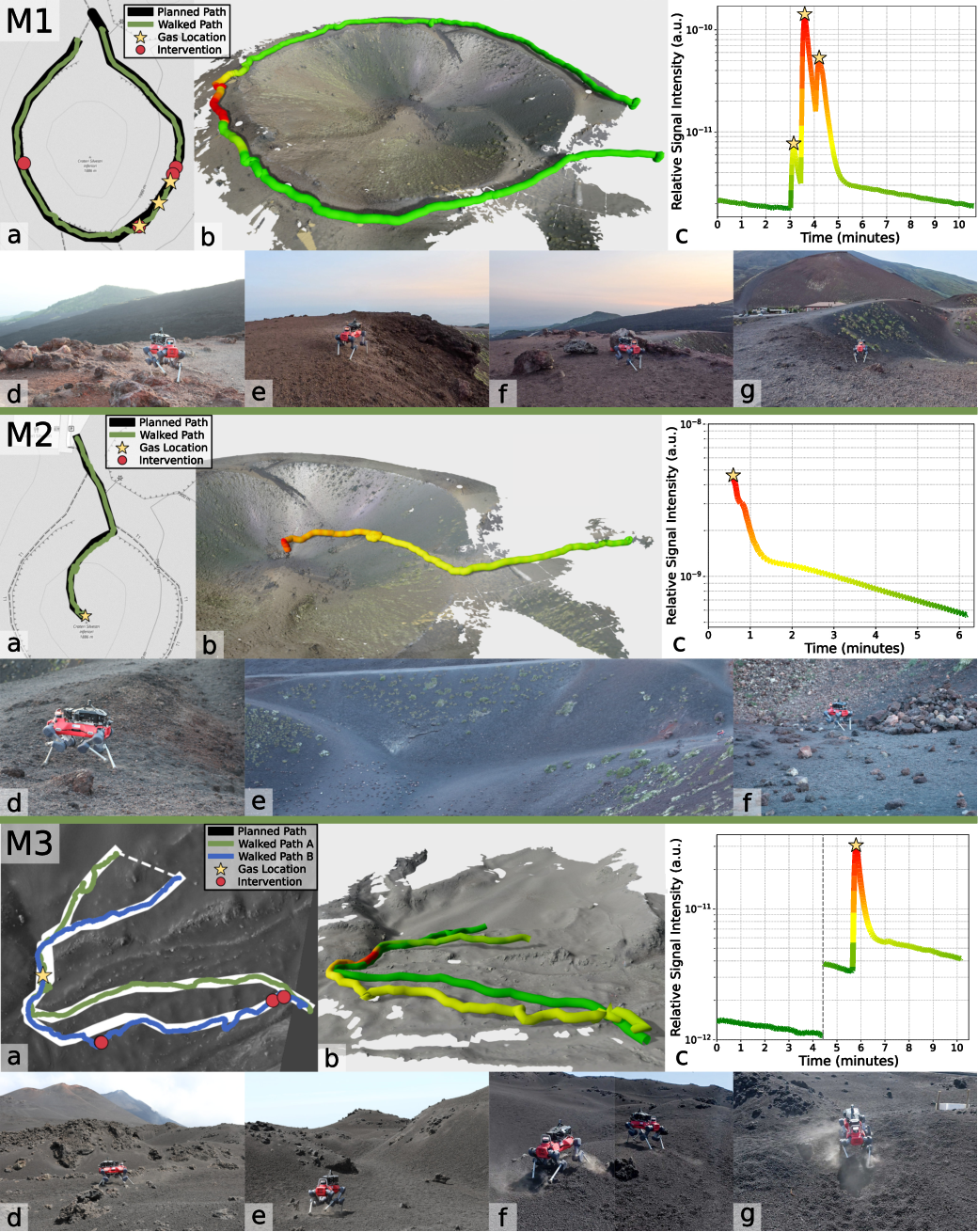}
    \caption{
Results of the three autonomous field missions on Mount Etna: M1-Crater Rim, M2-Crater Descent, and M3-Volcanic Desert.
For each mission: (a) Satellite image showing the mission area with marked \textcolor[HTML]{d43946}{interventions} and successful \textcolor[HTML]{e7ab00}{gas source detections}. (b) Terrain mesh with the traversed path, color-coded by relative gas concentration. (c) Time series of gas measurements using the same color map. (d--g) Representative images illustrating the robot's operation in varying terrain conditions.}
    \label{fig:mission_overview}
\end{figure*}
}

\subsection{Autonomous Mission Results}

The three autonomous missions were analyzed with respect to both gas-sensing performance and autonomous navigation reliability.
Table~\ref{tab:mission_metrics} summarizes the deployment results.
Two categories of performance measures were considered.
First, gas-source detection was assessed by verifying, through post-mission analysis, whether the deployed sources were successfully identified in the recorded concentration data.
For this, elevated readings from the mass spectrometer confirm that the robot has entered a region of active degassing. The observed concentration peak indicates proximity to a source, though wind-driven advection can displace these measurements from the exact vent location.
Second, autonomy was quantified using two complementary metrics.
The Robot Attention Demand (RAD)~\cite{olsen2003metrics} relates the average duration of operator interventions to the average interval between them, and the Autonomy Rate (AR) gives the fraction of total mission time during which the robot operated without human input. 
Both definitions are given in the supplementary material.

All three missions were completed successfully, achieving RAD values below 0.1 and autonomy rates above 90\%.
Mission~2 was executed entirely autonomously, requiring no operator intervention, whereas Missions~1 and~3 involved short interventions to ensure equipment safety.
Five of the eight deployed sources were successfully detected.
Since the robot followed a predefined path without adapting to the live concentration readings, the plume had to cross the inlet during the traversal. Average winds of \SI{6.2}{\kilo\meter\per\hour} and gusts up to \SI{39.2}{\kilo\meter\per\hour}, according to meteorological reports for the test days, dispersed the plumes and made detection sensitive to source placement.
% https://open-meteo.com/en/docs/historical-weather-api?start_date=2025-06-03&latitude=37.7223471003515&longitude=15.004117443002375&end_date=2025-06-05&hourly=temperature_2m,wind_speed_10m,wind_direction_10m,wind_gusts_10m#hourly_weather_variables
For each mission, Figure \ref{fig:mission_overview} shows the full trajectory projected onto a high-resolution mesh, the corresponding gas concentration time series, and representative images captured during deployment.

\textbf{Mission 1 – Crater Rim: }
The first experiment was conducted along the rim of the Silvestri crater.
Out of the five helium sources deployed, the robot detected three at the correct locations, visible as clear peaks in the recorded signal. Due to the wind along the crater rim, the remaining two sources were not detected, as the gas plumes were blown away from the robot.
For the identified detections, the time series in Figure~\ref{fig:mission_overview} (M1c) shows a steep rising edge and a slower decay, which is typical for mass spectrometers due to residual gas remaining in the vacuum chamber after exposure.

\begin{figure}[t]
    \centering
    \includegraphics[width=\linewidth]{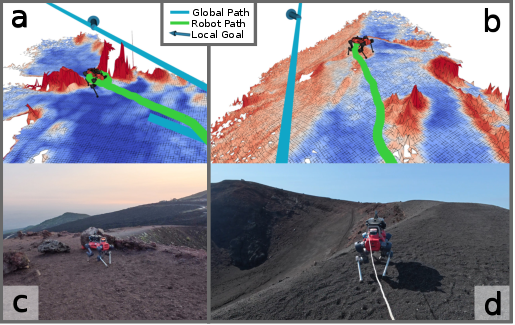}
    \caption{
Example of global-local path misalignment during an autonomous mission.
(a,b) Mission GUI showing the \textcolor[HTML]{10a6c8}{global path}, the traversable trail (\textcolor[HTML]{3a4cc0}{blue} regions in the elevation map), and the robot's \textcolor[HTML]{39ce3d}{past trajectory}.
(c,d) Corresponding camera view of the same scenes.}
    \label{fig:path_misalignment}
\end{figure}

Four operator interventions were required during this mission, as indicated in Figure~\ref{fig:mission_overview} (M1e--M1g).
These were caused by the local planner overestimating obstacle clearance (M1f), incorrect traversability estimation on fine ash (M1e), and a subtle misalignment of the robot heading in the world frame (M1g).
In all cases, the transformed global path deviated several meters from the real trail, as illustrated in Figure~\ref{fig:path_misalignment}.
The offset stemmed from limited OpenStreetMap positional accuracy combined with intermittent SLAM drift in weakly structured terrain, further amplified by GNSS uncertainty of approximately $\pm\SI[parse-numbers=false]{2-3}{\meter}$.
Because global orientation observability relies on both reliable SLAM geometry and an accurate GNSS position anchor, the simultaneous degradation of these two sources caused the fusion system to misinterpret small positional inconsistencies as changes in orientation, producing a rotational offset in the projected global path.
As a result, the lookahead waypoints were physically unreachable, leading the local planner to generate commands that steered the robot into local minima or low-traversability regions.

\textbf{Mission 2 – Crater Descent: }
For the second experiment, the robot descended into the Silvestri crater and returned to the rim.
The path was moderately steep with partially loose ground, and the crater floor consisted of a boulder field with small to medium-sized rocks (Figure~\ref{fig:mission_overview} (M2f)), both posing notable locomotion challenges.

The robot successfully descended into and exited the crater without any human intervention.
The main challenge arose from the rocks on the crater floor.
The elevation map struggled to capture these smaller obstacles, occasionally causing the robot to step on loose rocks, though the locomotion controller proved sufficiently robust to recover.
A single helium source was placed at the center of the crater.
To illustrate a case where gas measurements are taken only in regions of expected degassing, the mass spectrometer was only activated once the target area was reached. The resulting gas concentrations measured during the exit from the crater are shown in the concentration map in Figure~\ref{fig:mission_overview} (M2b).

\textbf{Mission 3 – Volcanic Desert: }
For this experiment, the robot traversed a volcanic desert near the Laghetto crater.
A particular challenge in this mission was the traversal of a narrow valley (Figure~\ref{fig:mission_overview} (M3e)).
The mission was planned from a road to a target location in the center of the field and back, as shown in Figure~\ref{fig:mission_overview} (M3a) in \textcolor[HTML]{759551}{green} and \textcolor[HTML]{4065c0}{blue}, respectively.
An untraversable hill between the start and goal required the global path to be routed around it, as the local planner could not account for an obstacle of that scale.

Three operator interventions were required: once when the locomotion controller failed to clear a small hole next to a rock (repeated foot collisions), and twice when insufficient forward velocity caused slipping in fine sand uphill (Figure~\ref{fig:mission_overview} M3g).
Figure~\ref{fig:mission_overview} (M3f) shows another incident, where the robot failed to climb on the left side of a rock but succeeded on the right during the second attempt, highlighting how identical slopes and appearances can mask different soil behaviors.
One challenge for localization occurred in the narrow valley, where GNSS uncertainty increased to approximately \SI[parse-numbers = false]{\pm6}{\meter}.
Despite the reduced signal quality, the consistent local frame provided by the SLAM system helped the fusion-based localization maintain a stable global pose estimate and prevent deviation with respect to the target global robot position. 
A helium source was placed inside the valley, shown as a star in Figure~\ref{fig:mission_overview} (M3a).
The robot did not detect the gas on its first pass, likely because the plume was dispersed by the wind away from the inlet, but it successfully measured it during the return traversal, as shown in the time series in Figure~\ref{fig:mission_overview} (M3c).
In this plot, the grey line indicates a measurement pause, during which the operator temporarily halted data collection and repositioned the robot to better analyze the surroundings.

\section{VOLCANIC PLUME MEASURING}
\label{sec:PLUME MEASURING}
\begin{figure*}[t]
    \centering
    \includegraphics[width=.95\textwidth]{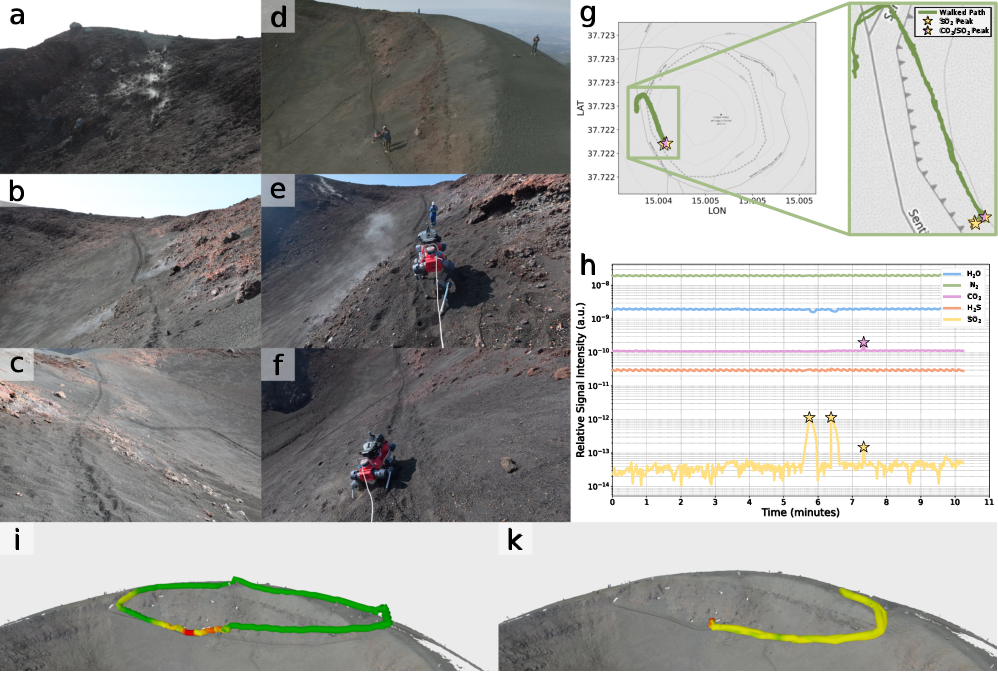}
    \caption{
Overview of the fumarole measurement on Mount Etna.
(a--c) Close-up views of fumaroles illustrating typical gas emission features.
(d--f) Photographs of the legged robot during operation.
(g) Satellite data showing the traversed path with highlighted gas peaks.
(h) Time series of measured gas concentrations recorded during the mission, with peaks indicated.
(i) Comparison measurement from the miniGAS instrument showing \ce{CO2} concentration.
(k) Terrain mesh with the traversed path, color-coded by relative \ce{SO2} concentration.}
    \label{fig:plume_viz}
\end{figure*}

We conducted one teleoperated test near active fumaroles inside the Laghetto crater on Mount Etna.
These fumaroles were located on a side slope with very loose ground, as shown in Figure~\ref{fig:plume_viz}a--f.
The path leading to the plumes was narrow and ran nearly parallel to the contour lines, with a subtle inclination relative to the horizontal ground.
While the central foot trail consisted of compacted soil, the surrounding areas were covered with loose, deep sand.
Autonomous operation was infeasible, as even small lateral steps caused the robot's feet to sink deeply into the substrate (Figure~\ref{fig:plume_viz}f).
The teleoperated mission took place under moderate wind conditions, which caused plume dilution and made gas measurements challenging, as the plume shifted rapidly around the robot.
We monitored water vapor (\ce{H2O}, \SI{18}{amu}), hydrogen sulfide (\ce{H2S}, \SI{34}{amu}), carbon dioxide (\ce{CO2}, \SI{44}{amu}), and sulfur dioxide (\ce{SO2}, \SI{64}{amu}) as key volcanic volatiles, and nitrogen (\ce{N2}, \SI{28}{amu}) as a background reference.
The robot was commanded to move along the slope, occasionally stopping in areas where water vapor was visible.\looseness-1

Figure~\ref{fig:plume_viz}k visualizes the measurement site and the robot's trajectory, while Figure~\ref{fig:plume_viz}h shows the recorded gas concentrations over time.
A distinct peak in sulfur dioxide is visible, corresponding to the time when the robot was directly adjacent to the fumarole. 
A smaller rise in sulfur dioxide and carbon dioxide was detected when the robot started its return.
In contrast, no clear increase in water vapor concentration was observed, likely due to the already high ambient humidity levels, which made plume-related variations difficult to distinguish.

To validate the onboard measurements, we manually collected reference readings using a handheld miniGAS instrument (INFICON Hapsite\textsuperscript{\textregistered} Scout), which includes electrochemical sensors for \ce{CO2} and \ce{SO2}.
The handheld instrument was carried along the same path and positioned in visible plumes to obtain reference readings.
Figure~\ref{fig:plume_viz}i shows the corresponding \ce{CO2} measurements plotted at the recorded locations, confirming the same fumarole positions as observed in Figure~\ref{fig:plume_viz}k.
While the miniGAS \ce{SO2} sensor did not register a clear signal, the robot-mounted mass spectrometer successfully detected a pronounced sulfur dioxide peak from approximately \SI{0.5}{\meter} away.
Conversely, the stronger \ce{CO2} response of the handheld sensor can be attributed to its direct proximity to the emission source.
Together, these results confirm the consistency between the two sensing methods and support the reliability of the robot-mounted mass spectrometer for in situ volcanic gas analysis.

\section{LESSONS LEARNED}
\label{sec:LESSONS LEARNED}
\subsection{Volcanic Gas Analysis}

The robot successfully traversed the volcanic terrain in order to reach active fumaroles and conducted gas measurements, detecting distinct gas compositions from different plumes.
However, the real-world deployment also revealed several practical challenges that are important to consider for the future.
First, volcanic gas plumes are highly transient, particularly under strong wind conditions.
Even with a fast and sensitive instrument such as the mass spectrometer, plume dispersion occasionally prevented accurate detection.
This highlights the need for measurement strategies that better account for environmental conditions, such as incorporating visual gas-source detection or wind sensing for plume prediction.
Since fumarolic emissions often contain water vapor visible as steam plumes, visual gas-source detection could enable the identification of active vents from a distance, allowing the robot to adjust its path toward a detected source.
Once an initial gas trace is encountered, techniques from the field of \ac{GSL} could further guide the robot toward the region of highest concentration.
However, as discussed in Section~\ref{sec:INTRODUCTION}, existing GSL methods are not validated for the wind conditions encountered on volcanic summits and would need to be adapted accordingly.

Second, for ground-level gases at low concentrations, the current inlet setup, mounted approximately \SI{0.75}{\meter} above the ground, may be too high to effectively sample the plume.
Future solutions would benefit from a deployable inlet mechanism capable of positioning the sampling point directly at or near the emission source. \looseness-1

\subsection{Autonomy and Mobility}

The autonomous missions demonstrated that reliable navigation on volcanic terrain is feasible, with autonomy rates exceeding 90\%.
Across three distinct environments (crater rim, crater descent, and volcanic desert), the robot consistently executed complex paths with minimal operator intervention, confirming the robustness of the navigation stack.
Nevertheless, the few required interventions provided valuable insights into the remaining limitations of the system and highlighted directions for future development.

\textbf{Planning} based on global satellite data remains challenging due to its coarse resolution, limited positional accuracy, and a lack of terrain detail necessary to accurately assess traversability a priori.
When following existing trails, the global path accuracy was occasionally affected by misalignments between the OpenStreetMap data and the actual ground path.
When planning across open terrain, the operator relied on prior knowledge of large obstacles that were not visible in satellite imagery, such as the untraversable hill in the volcanic desert.
Moreover, locally untraversable regions, such as slopes exceeding the robot's climbing capability, may only become apparent once the robot attempts to execute the global plan.
Although such offsets are expected, they can result in infeasible goals for the local planner.
In the current system, such cases are not handled autonomously and require operator intervention. 
An exploration planner could mitigate this issue by finding alternative paths around obstacles. However, for large obstacles such as the untraversable hill encountered in Mission~3 (Volcanic Desert), this approach can result in unnecessarily long or inefficient detours, thereby reducing the robot's overall operational range during long missions.
To overcome these limitations, future systems should reduce the information gap between global and local planning.
Rather than propagating only discrete waypoints, we propose continuously integrating local observations and feeding them back into the global map to update the current path.
This process could be complemented by active perception, where the robot scans its surroundings to evaluate multiple feasible route options before proceeding.\looseness-1

\textbf{Perception and Localization} proved challenging in the volcanic environment.
The robot's motion stirred up dust and fine sand, which intermittently degraded both depth camera and LiDAR data.
LiDAR-based localization also suffered: scan matching frequently became underconstrained on planar, self-similar slopes, and transient dust returns introduced noise points that increased alignment error and drift.
While GNSS provides global positioning, satellite coverage was often poor near crater rims and steep walls, resulting in a low-quality signal with high variance and occasional dropouts. On the other hand, LiDAR provides accurate but feature-deprived measurements in similar environments.
Some alternatives to complement these sensors in such environments include radar, Frequency-Modulated Continuous-Wave LiDAR, and thermal cameras, which can provide reliable measurements across different spectra.
Moreover, visual appearance often did not reflect the true soil stability, leading to errors in traversability estimation.
A promising solution is adaptive traversability estimation, in which the robot refines its terrain model using feedback from ground interactions.

\textbf{Locomotion} performance during the field tests demonstrated robust operation on various types of volcanic terrain, examples of which can be seen in the supplementary video.
Over five days of testing across diverse surface types, the robot experienced no falls, highlighting the maturity of current quadruped locomotion policies.
Nevertheless, when confronted with even more demanding conditions, such as those encountered during the teleoperated tests, limitations became apparent.
Achieving full autonomy capable of reaching fumaroles will likely require reliable traversal of very fine sand on steep slopes – conditions that remain challenging for existing foot designs and locomotion strategies.
Improved foot geometries and adaptive contact mechanisms are promising directions for increasing stability and traction under these conditions.

\section{CONCLUSION}
\label{sec:CONCLUSION}
In this work, we present a robotic system for autonomous measurement of volcanic gases in challenging terrain. 
The design integrates a mass spectrometer for fast and accurate in~situ gas analysis with an autonomy stack that enables long-range traverses across unstructured terrain.
Through a series of field trials on Mount Etna, we demonstrated robust navigation across complex volcanic terrain and performed the first successful in~situ gas measurements by a ground-based mobile robot on an active volcano, serving as a proof of concept. 
These results establish the proposed system as a viable solution for autonomous volcanic gas analysis, highlighting its potential for advancing field volcanology.

%%%%%%%%%%%%%%%%%%%%%%%%%%%%%%%%%%%%%%%%%%%%%%%%%%%%%%%%%%%%%%%%%%%%%%%%%%%%%%%%
\ifthenelse{\boolean{anonymous}}%
{} % Show no acknowledgments
{ % Show acknowledgments
\section*{ACKNOWLEDGMENT}
The authors would like to thank Simona Scollo, Fabrizia Buongiorno, Gaetano Giudice, Roberto Maugeri, Emilio Pecora, and Luca Merucci from the Istituto Nazionale di Geofisica e Vulcanologia (INGV) for making the Mount Etna field campaign possible, including support with permits and logistical coordination for transport onto the volcano. We further thank Dario Calogero Guastella and Giuseppe Sutera from the University of Catania (ROSYS Group) for their valuable on-site support during the Etna experiments.

This work was supported by INFICON Inc., the Luxembourg National Research Fund (Ref. 18990533), and the Swiss National Science Foundation (SNSF) as part of the projects No. 200021E\_229503 and No. 227617.
}

%%%%%%%%%%%%%%%%%%%%%%%%%%%%%%%%%%%%%%%%%%%%%%%%%%%%%%%%%%%%%%%%%%%%%%%%%%%%%%%%
\begingroup
\raggedright
\tiny
\bibliographystyle{IEEEtran}
\bibliography{bibliography}
\endgroup

%%%%%%%%%%%%%%%%%%%%%%%%%%%%%%%%%%%%%%%%%%%%%%%%%%%%%%%%%%%%%%%%%%%%%%%%%%%%%%%%
\clearpage
\ifshowappendix
  \appendices
  % Appendix: secondary technical details moved out of the main text
% to satisfy the RA-M page limit. Intended for the supplementary material.

\section*{Appendix}
\subsection{Factor-Graph Formulation of Holistic Fusion}
\label{app:map}

As described in Section~\ref{subsec:state_estimation} the utilized state estimator builds on \ac{HF}~\cite{nubert2026holistic}, which formulates estimation as a factor-graph \ac{MAP} problem
\begin{equation}
  \states^\star \;=\; \argmax_{\states}\, p(\states \mid \Measurements)
  \;\propto\; p(\Measurements \mid \states)\, p(\states),
  \label{eq:map}
\end{equation}
where \(\Measurements\) collects all sensor measurements and priors, and \(\states\) comprises the robot and alignment variables
\begin{equation}
  \states \;=\; \{\, {}^{I}\states_{N_I},\; {}^{G}\states_{N_G},\; {}^{R}\states_{N_R} \,\}.
  \label{eq:states}
\end{equation}
Here, \({}^{I}\states_{N_I}\) denotes the time-indexed robot state in the world frame, while \({}^{G}\states_{N_G}\) and \({}^{R}\states_{N_R}\) parameterize the global and reference-frame alignments that couple the local map with the global reference frame. The two design choices introduced in Section~\ref{subsec:state_estimation}, namely absolute \ac{LiDAR}-SLAM pose factors and explicit estimation of the map-to-world alignment, act on \({}^{R}\states_{N_R}\) and on the corresponding factors in this graph. This makes the factor graph robust to drifts in LiDAR-SLAM and misalignments arising from measurement uncertainties.

\subsection{Curvature-Adaptive Lookahead Sampling}
\label{app:lookahead}

This appendix gives the full explanation of the lookahead sampling scheme summarized in Section~\ref{subsec:local_nav}. The procedure is illustrated in Figure~\ref{fig:path_curvature}.

The robot position \(\mathbf{x}_{\mathrm{robot}}\) is first projected onto the global path to obtain the closest anchor point \(\mathbf{x}_i\) at index \(i\). Two secant vectors \(\mathbf{v}^{(p)}_i\) and \(\mathbf{v}^{(n)}_i\) are then constructed, pointing forward and backward along the path from the anchor. Their normalized directions \(\hat{\mathbf{v}}^{(p)}_i\) and \(\hat{\mathbf{v}}^{(n)}_i\) define the local curvature \(\kappa_i\) as
\begin{equation}
    \kappa_i = \frac{\theta_i}{\lVert \hat{\mathbf{v}}^{(n)}_i \rVert},
    \qquad
    \text{with } \theta_i = \arccos\!\big(\hat{\mathbf{v}}^{(p)}_i \cdot \hat{\mathbf{v}}^{(n)}_i\big),
    \label{eq:curvature}
\end{equation}
where \(\theta_i\) is the angular difference between the two vectors. The adaptive lookahead distance follows as
\begin{equation}
    L(\kappa_i) = L_{\min} + \frac{L_{\max} - L_{\min}}{1 + \lvert \kappa_i \rvert / \kappa_{\mathrm{ref}}},
    \label{eq:lookahead}
\end{equation}
with \(L_{\min}\) and \(L_{\max}\) the minimum and maximum lookahead distances, and \(\kappa_{\mathrm{ref}}\) controlling the sensitivity to curvature changes. Equation~\eqref{eq:lookahead} decreases monotonically in \(\lvert \kappa_i \rvert\), so that straight path segments yield a long lookahead and tight turns a short one.

Given \(L(\kappa_i)\), we place \(N\) equally spaced samples along the path ahead of the anchor point. The lookahead waypoint \(\mathbf{x}_{\mathrm{goal}}\) is obtained as an exponentially weighted average of these samples with respect to their arc-length distance from the anchor, which emphasizes the nearby path geometry while still accounting for the structure of the upcoming trajectory.

\begin{figure}[t]
    \centering
    \includegraphics[width=.9\linewidth]{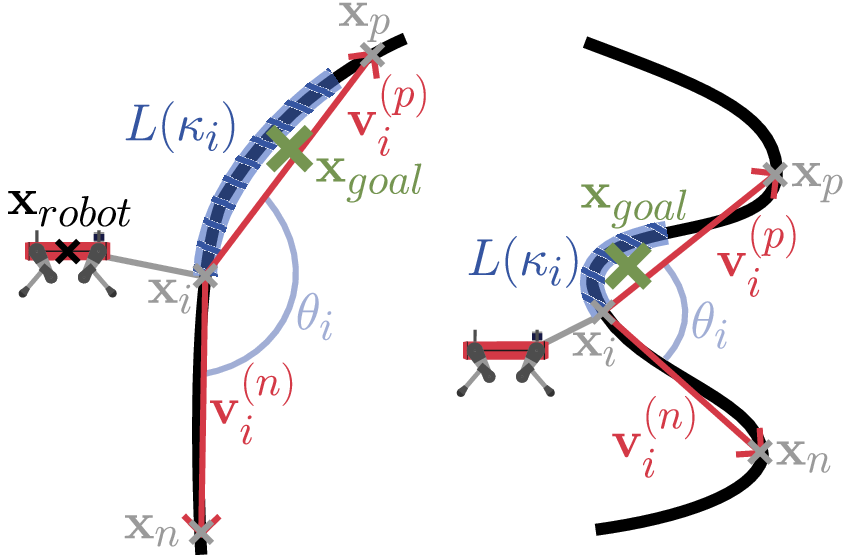}
    \caption{Computation of the curvature-based adaptive lookahead.
The robot position (\(\mathbf{x}_{robot}\)) is projected onto the path to obtain the nearest anchor point (\textcolor[HTML]{9a9a9a}{\(\mathbf{x}_i\)}), from which forward and backward secant vectors \big(\textcolor[HTML]{d43946}{\(\mathbf{v}^{(p)}_i\)}, \textcolor[HTML]{d43946}{\(\mathbf{v}^{(n)}_i\)}\big) define the local curvature (\textcolor[HTML]{3555a1}{\(\kappa_i\)}).
The lookahead waypoint (\textcolor[HTML]{759551}{\(\mathbf{x}_{\mathrm{goal}}\)}) is then placed according to the curvature-dependent lookahead distance (\textcolor[HTML]{3555a1}{\(L(\kappa_i)\)}).}
    \label{fig:path_curvature}
\end{figure}

\subsection{Autonomy Metrics}
\label{app:metrics}

The two autonomy metrics reported in Table~\ref{tab:mission_metrics} are defined as follows. The Robot Attention Demand~\cite{olsen2003metrics} is
\begin{equation}
  \mathrm{RAD} = \frac{\mathit{IE}}{\mathit{IE} + \mathit{NT}},
  \label{eq:rad}
\end{equation}
where the interaction effort \(\mathit{IE}\) denotes the average duration of operator interventions and the neglect tolerance \(\mathit{NT}\) the average interval between them. Lower values indicate that the robot required less operator attention per unit of autonomous operation. The Autonomy Rate expresses the fraction of total mission time during which the robot operated without human input,
\begin{equation}
  \mathrm{AR} = 100 \times \left(1 - \frac{T_{\mathrm{intervention}}}{T_{\mathrm{total}}}\right),
  \label{eq:ar}
\end{equation}
given in percentage.
\fi

\end{document}